\documentclass[runningheads]{llncs}
\usepackage[
 activate={true,nocompatibility}, % activate protrusion and expansion
 final, 			% enable microtype; use "draft" to disable
 tracking=true,
 kerning=true,
 spacing=true,
 factor=1100, 	% add 10% to the protrusion amount (default is 1000)
 stretch=20, 	% allow up to 2% stretch
 shrink=20 	% allow up to 2% shrinkage
]{microtype}

\usepackage[T1]{fontenc}
\usepackage{graphicx}

\usepackage{booktabs}
\usepackage{multirow}
\usepackage{xcolor}
\usepackage{caption}

\begin{document}
\title{Generating Planning Feedback for Open-Ended Programming Exercises with LLMs}

\titlerunning{Generating Planning Feedback in Programming with LLMs}
% \author{Author 1}
% \email{email1@email.edu}
% \orcid{}
% \affiliation{%
%   \institution{Institution}
%   \city{City}
% \title{Detecting Programming Plans in Open-ended Code Submissions}
%   \country{Country}
% }

% \author{Author 2}
% \email{email2@email.edu}
% \orcid{}
% \affiliation{%
%   \institution{Institution}
%   \city{City}
%   \state{State}
%   \country{Country}
% }

% \author{Author 3}
% \email{email3@email.edu}
% \orcid{}
% \affiliation{%
%   \institution{Institution}

%   \state{State}
%   \country{Country}
% }
\author{Mehmet Arif Demirta\c{s}\inst{1} \and
Claire Zheng\inst{1} \and
Max Fowler\inst{1} \and
Kathryn Cunningham\inst{1}}

% \author{Mehmet Arif Demirtas\inst{1}\orcidID{0000-1111-2222-3333} \and
% Third Author\inst{3}\orcidID{2222--3333-4444-5555} \and
% Fourth Author\inst{3}\orcidID{2222--3333-4444-5555}}
% %
\authorrunning{Demirta\c{s} et al.}
\institute{University of Illinois Urbana-Champaign, Urbana IL 61801, USA
\email{\{mad16,clairez5,mfowler5,katcun\}@illinois.edu}}
\maketitle              % typeset the header of the contribution
\begin{abstract}
To complete an open-ended programming exercise, students need to both plan a high-level solution and implement it using the appropriate syntax.
However, these problems are often autograded on the correctness of the final submission through test cases, and students cannot get feedback on their planning process.
Large language models (LLM) may be able to generate this feedback by detecting the overall code structure even for submissions with syntax errors.
To this end, we propose an approach that detects which high-level goals and patterns (i.e. \textit{programming plans}) exist in a student program with LLMs.
We show that both the full GPT-4o model and a small variant (GPT-4o-mini) can detect these plans with remarkable accuracy, outperforming baselines inspired by conventional approaches to code analysis.
We further show that the smaller, cost-effective variant (GPT-4o-mini) achieves results on par with state-of-the-art (GPT-4o) after fine-tuning, creating promising implications for smaller models for real-time grading.
These smaller models can be incorporated into autograders for open-ended code-writing exercises to provide feedback for students' implicit planning skills, even when their program is syntactically incorrect.
Furthermore, LLMs may be useful in providing feedback for problems in other domains where students start with a set of high-level solution steps and iteratively compute the output, such as math and physics problems.\footnote{Accepted at AIED 2025}

\end{abstract}

\section{Introduction}

%Code-writing requires operating at multiple levels of understanding, which can be overwhelming for students [mention process]

% Programming is a complex task that involves planning and implementation.
% Common activities are code writing problems graded by test bcases
% Test cases only provide insight baout the outcome of the code, the correctness of the final code. There is no direct feedback about the **process** the learner should take to get to the code. 

Learning programming requires applying several skills at the same time, which may be overwhelming for novice students~\cite{robinsNoviceProgrammersIntroductory2019}.
% Learning programming requires multiple skills at once, such as forming mental models of computing concepts, utilizing problem-solving strategies, and applying syntax knowledge~\cite{robinsNoviceProgrammersIntroductory2019}.
To solve a problem, a student will likely \textit{interpret} a problem statement, \textit{decompose} it into smaller chunks, \textit{plan} how these chunks will connect, and \textit{implement} it using the correct syntax. 
In psychology of programming literature, the chunks in this planning process have been termed `programming plans': common code snippets that have a clear goal~\cite{soloway1986learning}. Experts in programming recall and implement these plans as they decompose a problem for efficient problem-solving~\cite{solowayEmpiricalStudiesProgramming1984}. Implicit or explicit instruction about plans is a common approach to support students' problem-solving processes~\cite{iyerPatternCensusCharacterization2021,demirtacs2024validating}. 

Programming courses commonly include open-ended code writing exercises to practice this end-to-end process of planning and implementation.
% measures many of these skills at the same time: the student needs to interpret a problem statement using their mental models, select the correct strategy for the given problem, and implement the strategy in a particular language syntax.
Moreover, these exercises can be autograded with test cases, making them easily scalable to hundreds of students.
However, test-case based autograders do not necessarily identify the source of errors, whether it is starting with the wrong plan for the problem or making a small syntax error in the final submission.
With autograders that rely on test cases, students only see the \emph{outcome} of the executed code, and they do not get feedback for the skills required in the \emph{process} for solving programming problems.
If we can develop autograders to detect how a student has decomposed a problem, \textit{independent of how well they implemented their solution}, we can provide feedback on their planning skills.

% An ideal autograder should provide additional feedback on the planning skills of the student to highlight room for improving planning skills.%which competencies students need to practice and improve.

% whether their code fails due to a conceptual mistake at the planning stake or due to a simple typo.
% For example, a student might start with an inefficient template for a problem due to an inaccurate mental model, and get frustrated as they try to debug their code.
% Alternatively, they might get a problem wrong due to a simple syntax error, even though they completed most of the work on understanding the problem statement and translating it into an algorithm.
% these autograders treat all student mistakes equal, whether it is starting with the wrong template for the problem in the planning stage or making a small syntax error in the final submission. Moreover, instructors may not be able to intervene for students who write code to pass test cases through inefficient problem solving strategies.

Prior literature on large language models (LLM) for programming suggest that LLMs can interpret code and extract high-level semantic information~\cite{alkafaweenAutomatingAutogradingLarge2024,smithPromptingComprehensionExploring2024,duongAutomaticGradingShort2024,grandelApplyingLargeLanguage2024,yousefBeGradingLargeLanguage2024,dongGradingProgrammingAssignments2024}, making them appropriate for inferring intermediate planning skills from a final program. 
However, LLMs may generate hallucinations, where the output is incorrect or inaccurate in subtle ways~\cite{Ji_2023_hallucination}.
To prevent hallucinations in a specialized domain, some researchers have suggested constraining the LLM to generate \textit{structured output} within a framework that is informed by that domain~\cite{xieyang2024structuredoutput}.
% One solution to reduce hallucinations is to constrain LLM outputs to a specific structure using frameworks from a domain~\cite{xieyang2024structuredoutput}.
% We propose leveraging programming plans literature from computer science education as a framework for classifying common planning activities in open-ended programming exercises. 
To this end, programming plan literature from computing education research can constitute a framework for understanding common planning activities in open-ended programming exercises. 

% Thus, we can provide feedback on a student's planning behavior by classifying their submission with a set of high-level programming plans, and comparing the detected plans to the plans that are expected for that problem using the instructor solution.
% regardless of the test case correctness.
% Due to large corpora of code these models are trained on, they can be utilized to identify common patterns in student programs and provide feedback on...

In this work, we utilize large language models (LLMs) to provide feedback on planning skills based on a student's final code submission, regardless of test case correctness. Specifically, we formulate a classification task where the student submission to an open-ended code-writing exercise is classified into a set of predefined programming plans to answer the following research question:

    \textbf{RQ:} To what extent can a student's \textit{intended} programming plan be detected from their (possibly incorrect) code submission in introductory programming?

% We first manually annotate data from student submissions in a CS1 course, showing that mismatches between instructor and student plans as a notable factor for failing submissions. Then, we evaluate a state-of-the-art LLM (GPT-4o~\cite{}) and a smaller, cost-effective version (GPT-4o-mini~\cite{})
% through both few-shot prompting and fine-tuning.
% We compare these models against baselines using rules from static analysis of the submissions using ASTs, and using semantic embeddings obtained using CodeBERT~\cite{}. 

% Through an evaluation on student submissions from a CS1 course, we show that LLMs (GPT-4o~\cite{openai2024gpt4ocard} and GPT-4o-mini~\cite{gpt4omini}) are remarkably better at plan classification compared to traditional methods for code analysis, even with no fine-tuning.
% More importantly, we find that fine-tuning GPT-4o-mini with a small dataset can improve performance significantly, with fine-tuned GPT-4o-mini achieving a higher micro-F1 score at (.781) than GPT-4o (.778). 
To answer this RQ, we evaluate our models on student submissions from a CS1 course. Our findings suggest that LLMs enable feedback on implicit skills in programming, and small and cost-effective models can be used to provide feedback on programming plans in open-ended code-writing exercises. 

\section{Related Works}

\subsection{Programming Plans}

Although programming requires thinking at varying levels of granularity, a core cognitive unit in programming problem-solving processes is identified as the programming plan~\cite{robinsNoviceProgrammersIntroductory2019}. A programming plan is a common code pattern for achieving a goal, such as counting items in a list. Structures similar to plans have been reviewed under many names including ``programming patterns''~\cite{iyerPatternCensusCharacterization2021}, ``templates''~\cite{clancy_designing_1996}, ``algorithmic patterns''~\cite{mullerPatternOrientedInstruction2005}, and ``plan-schemata''~\cite{kather2021through}.
The ability to notice these underlying patterns appears to be a key part of programming expertise~\cite{solowayEmpiricalStudiesProgramming1984}.
It has also been shown that experts' recall of plan structures is associated with being more efficient at problem-solving compared to novices~\cite{ristProgramStructureDesign1995}.

Programming plans are taught in many introductory CS courses, either explicitly or implicitly~\cite{iyerPatternCensusCharacterization2021,xieTheoryInstructionIntroductory2019,mullerPatternorientedInstructionIts2007}. Evidence from student submissions in CS1 courses suggests that novice learners improve at applying these plans with more practice~\cite{demirtacs2024validating}. There have been several instructional interventions leveraging plan-like structures to support students in planning stages, without necessarily requiring code-writing~\cite{Weinman_Fox_Faded_Parsons,Cunningham_PurposeFirstProgramming_CHI-2021,reichertJigsawToolDecomposing2024,riveraIterativeStudentProgram2024}.
For instance, Jigsaw~\cite{reichertJigsawToolDecomposing2024} presented students with a set of programming plans to help them compose their solution to practice planning skills before writing code. Rivera et al.~\cite{riveraIterativeStudentProgram2024} designed a planning workflow that asks students to describe their solution in natural language. They evaluated the workflows by using LLMs to generate code based on these descriptions but reported difficulties in providing high-level feedback with this approach due to the quality and structure of the LLM outputs.
Moreover, when planning activities are distinct from code-writing exercises, some students may find them less authentic or frustrating~\cite{riveraObservationsDesignProgram2024a}.
Thus, detecting a student's planning logic from the output of a code-writing exercise can help students practice programming plan knowledge without losing the benefits of open-ended code-writing exercises.

\subsection{Detecting Structures in Code}
\label{sec:rw-baselines}
% Detecting which programming plans students are attempting to use in their submission is a first step towards inferring their planning process.

While the problem of inferring plan usage from code submissions has not been directly addressed, problems related to detecting structures and patterns in code have been explored in different contexts.
For instance, software engineering and artificial intelligence communities have explored mining \textit{code idioms}: a piece of code that has a semantic purpose and appears across projects~\cite{haggis_code_similarity}. These studies focused on identifying emergent patterns in large codebases rather than using a pedagogically verified set of patterns~\cite{haggis_code_similarity,jezero_code_similarity,code_similarity_newest,shin2019idiomssynthesis}. Thus, their utility in educational contexts is limited. 
An alternative approach was using canonical graph representations to recognize a set of patterns in programs~\cite{willsAutomatedProgramRecognition1987}. While this approach identifies coding patterns that are similar to programming plans, the system was limited to functional languages and could not process some data abstractions. We hypothesize that LLMs can avoid these limitations as they are trained on large corpora of code obtained from many languages.

A similar task in computing education research is detecting subgoals for problems. Similar to programming plans, subgoals break down larger problems into chunks with clear objectives. Providing formative feedback on subgoals can improve student motivation and support problem-solving skills~\cite{marwanJustFewExpert2021}. However, expert-authored feedback for student progression is unfeasible due to the large number of possible solutions~\cite{toll2020current}. To address this, Marwan et al. proposed a data-driven approach with expert constraints to identify subgoals a student intends to implement as they are writing code in real-time~\cite{marwanJustFewExpert2021}. However, their approach identifies subgoals on a problem basis and requires a set of submissions on the given problem for the initial training. In contrast, we identify problem-agnostic programming plans that can be extended to new problems with no additional data, supporting the generalizability of the autograder to new problems.

\subsection{Autograding Programming Problems with LLMs}
LLMs have created promising opportunities for autograding open-ended assessment items, as they achieve remarkable results even with no fine-tuning on additional data~\cite{wuMatchingExemplarNext2023,chenMultitaskAutomatedAssessment2024}. These models have proven to be particularly useful in programming domains, potentially due to being trained on a large number of open-source projects available online~\cite{finnie-ansleyRobotsAreComing2022,finnie-ansleyMyAIWants2023,wangExploringRoleAI2023}.
Studies have shown that LLMs can explain code more accurately than students~\cite{leinonenComparingCodeExplanations2023} and describe code on different levels of abstraction for learners~\cite{juryEvaluatingLLMgeneratedWorked2024a}. Furthermore, autograders have incorporated LLMs for various purposes: generating test cases for programming exercises~\cite{alkafaweenAutomatingAutogradingLarge2024}, evaluating short answer questions~\cite{smithPromptingComprehensionExploring2024,duongAutomaticGradingShort2024}, applying rubrics on student assignments~\cite{grandelApplyingLargeLanguage2024,yousefBeGradingLargeLanguage2024}, and summarizing student code~\cite{dongGradingProgrammingAssignments2024}. Thus, LLMs might be appropriate for inferring high-level structure and underlying patterns from student submissions to detect programming plans.

\section{Methods} 

We formulate the plan detection task as a classification problem where student submissions are classified as containing zero or more programming plans. We evaluate the models against human labels on 1616 student submissions. In this section, we describe our dataset, our baseline approaches, our approaches using LLMs, and an ablation study for testing the robustness of our methods.

\subsection{Dataset} 
\label{sec:dataset}

Our dataset is collected in an introductory Python course at a large public US university with IRB approval. The course data was collected over 7 semesters between 2019 and 2022, with the course primarily serving first and second-year undergraduate students from Business and Liberal Arts and Sciences majors. 42.6\% of students in the class identified as female. 41.8\% of students were White, 21.3\% were Asian, 16.6\% were International, 11.9\% were Hispanic, 4.4\% were Black/African American, 3.2\% were Multi-race, and 0.9\% did not report.

The dataset contains 116 short coding homework problems, one instructor-written solution per problem featuring at least one programming plan (annotated by~\cite{demirtacs2024validating}), and up to 30 student submissions per problem. %Figure \ref{fig:example-pl} shows an example problem in the testing environment from the course.

For each problem, we classified student submissions into four categories: completely correct (i.e. passing all test cases), partially correct (i.e. passing at least one test case), semantically incorrect (i.e. passing no test cases), and syntactically incorrect (i.e. failing at compilation).
For the first two categories, we started by collecting 10 submissions per problem. 
To capture diverse implementations in this process, we computed the abstract syntax trees (AST) for each submission to filter out structurally identical submissions. %but include surface-level differences.
For the latter two categories, we collected 3 submissions per problem and did not apply any filtering, resulting in 1616 submissions total.
% We sorted these ASTs based on frequency to select submissions corresponding to most common unique ASTs in problems. However, this also resulted in having less than 10 submissions with different ASTs for some problems.

\begin{table}[t]
    \centering
     \caption{Programming plans and their goals, adapted from \cite{iyerPatternCensusCharacterization2021}}
    \begin{tabular}{@{}lp{.64\columnwidth}@{}}
      \toprule
      \textbf{Plan} & \textbf{Goal} \\ 
      \midrule
      \texttt{processAllItems}         & Iterate over the items in a collection \\ 
      \texttt{filterACollection}         & Select items from a collection that satisfy a condition \\ 
      \texttt{findBestInCollection}      & Find the value that has the greatest value by an arbitrary measure \\ 
      \texttt{sum}                       & Compute the total of items from a collection \\ 
      \texttt{evennessCheck}             & Check if a number is even or odd using the modulus operator \\ 
      \texttt{counting}                  & Compute the number of items from a collection \\ 
      \texttt{booleanOperatorChaining}   & Combine multiple logic expressions to make a decision \\ 
      \texttt{multiWayBranching}         & Split into three different branches based on a logic expression \\ 
      \texttt{linearSearching}           & Find the first matching item that satisfies a condition \\ 
      \bottomrule
    \end{tabular}
   
    \label{table:3.1.2-plans}
\end{table}

Each student submission was annotated by the plan(s) included in the submission in an iterative process, using the list adapted from \cite{iyerPatternCensusCharacterization2021} (Table \ref{table:3.1.2-plans}). A submission might also be labeled as including multiple plans, or none of the known plans (denoted as \textit{UNKNOWN}).
A codebook was iteratively developed during the annotation process by the first two authors. %We used percent agreement to measure inter-rater reliability (IRR) as we are using multiple labels per item.
Initially, the first author and second author achieved a percent agreement of 90\% on completely correct submissions and 66\% on partially correct submissions after independently annotating 50 examples of each category. 
After reconciling disagreements through discussion and refining the codebook, both authors annotated 50 more examples in each of the four submission categories independently, achieving  93.9\% agreement.
% and 95.9\%.
% of 93.6\% and 95.9\% respectively (94\% on initial coding with refined codebook).

% \begin{verbatim}
% def count_strings_containing_substring(str_list, substr):
%     x = 0 
%     for str in str_list: 
%         if substr in str:
%             x += 1
%     return x
% \end{verbatim}

% def count_strings_containing_substring(str_list, substr):
%     x=0
%     for substr in str_list:
%         if substr in str_list:
%             x+=1
%     return x

% includes 116 problems with at least one plan - instructor solutions
% includes up to 10 unique correct and 10 partially correct
% mention the course context 
% maybe an example program?
%   - one true one false submission that are simliar enough to compare
% filtered for ambiguous solution 
    % solutions that have the same plan as the instructor solution
    % also have solutions that match other plans but we exclude this data in the dataset (extra and prolly not necessary to include)
    % or solutions that do not fit any plan
% have x and y number of correct and incorrect solutions (can use variables, arif will replace x and y with real numbers)

% We use the set of plans from ~Demirtas et al, which is a subset of Iyer plans validated on student data.

% We use a set of student submissions from CS105, a non major course. 
%     - Submissions that have full credit
%     - Submissions that have partial credit

\subsection{Baseline Approaches}

As prior work did not provide a directly applicable method for detecting a predefined set of plans across many problems (Sec \ref{sec:rw-baselines}), we propose two baselines to compare the performance of large language models against: \textit{AST-Rules} and \textit{CodeBERT-kNN}.

\textbf{Baseline 1: AST-Rules.}
We designed a rule-based classifier using abstract syntax trees (\textbf{ASTs}) of student submissions, inspired by \cite{willsAutomatedProgramRecognition1987}.
% The AST of a submission can be traversed to detect syntax structures (e.g. incrementing a counter) automatically, which can correspond to plans (e.g. counting).
An AST is a tree representation of the student code where each syntax element corresponds to a node in the tree. For example, an idea such as `increment variable inside for loop' can be represented as a subtree in the AST. Thus, by traversing the AST for a student submission with the right set of rules, we can detect whether or not a programming plan is implemented in the code.
Thus, we formulated a set of syntax structures (\textit{rules}) that are associated with each plan by a manual review of instructor solutions. Then, we implemented a rule-based classifier that checks these structures in the AST and detects the plans whose rules are satisfied. 
While this approach is less computationally expensive compared to LLMs, making it more feasible for autograding systems at scale, it also requires more instructor effort to identify rules. Moreover, it is more sensitive to small errors. For instance, incorrect indentation can lead to a drastically different AST. 

\textbf{Baseline 2: CodeBERT-kNN.}
We designed a k-Nearest-Neighbors (kNN) classifier using the code embeddings generated by CodeBERT model~\cite{feng2020codebert}. CodeBERT is a transformer architecture that creates numerical embeddings for representing code snippets. It has been shown to excel at tasks that require a high-level understanding of code, such as code search from natural language descriptions, and was the leading choice for code understanding tasks prior to the introduction of LLMs.
% It has been originally shown to excel at tasks that require high-level understanding of code, such as code search from natural language descriptions prior to LLMs.
We used CodeBERT to generate embeddings for both instructor solutions and student submissions. Then, we used a kNN classifier (k=3), classifying each student submission by comparing the labels of the instructor solutions with the most similar embeddings.

This approach is similar to that of LLMs as it relies on pretrained code embeddings, but it employs a more interpretable classification step due to comparing embeddings of code snippets directly. 
However, these embeddings may be sensitive to variable names and other surface-level details from the program, rather than summarizing the high-level structure of the plan the student used.

\subsection{LLM Approaches}

We used 2 models: GPT-4o~\cite{openai2024gpt4ocard} and GPT-4o-mini.
GPT-4o is a state-of-the-art model for code generation in Python~\cite{liu2023isyourcode}, whereas
GPT-4o-mini\footnote{https://openai.com/index/gpt-4o-mini-advancing-cost-efficient-intelligence/} is a smaller and more cost-effective variant. However, GPT-4o-mini may be more feasible to deploy at a large scale due to its size.
We evaluated these models in two settings: \textit{few-shot prompting} and \textit{fine-tuning}. 

\textbf{Few-shot Prompting.}
In few-shot prompting, examples of the target task are provided to the pretrained LLM as part of the prompt~\cite{brown2020languagemodelsfewshotlearners}.
Our prompt defined the nine relevant plans with three examples each and introduced the \textit{UNKNOWN} category for solutions where no known plan could apply. These examples are from the instructor solution set to prevent data leakage from student submissions.\footnote{Full prompts: https://github.com/marifdemirtas/AIED2025-Planning-Feedback}

\textbf{Fine-tuning.}
Fine-tuning is a technique for improving the performance of a pretrained LLM by further training it on examples from a particular task. As the pretrained models have already learned efficient embeddings, prior work has shown that a small number of examples can be sufficient to fine-tune these models and alter their behavior significantly, e.g. for automated scoring of constructed response problems~\cite{latif2024fine}. We fine-tuned both models on the dataset of 116 instructor solutions, for three epochs over the full dataset with a batch size of two at the same learning rate as the pretraining. For fine-tuned models, we used an alternative prompt for classification that does not include the few-shot examples.

\subsection{Ablation Study}
In addition to our full classification results, we conducted an \textit{ablation study} by replacing variable names and function signatures in programs with non-descriptive identifiers (e.g. var1) to evaluate the robustness of our models. We designed this ablation study to assess whether the models could infer structure from the code, or they were only using contextual cues (such as detecting the `counting' plan when there is a variable called `counter'). As novice students may not be using best practices when writing programs, we believe that this might result in a more realistic setting for our models.

For this study, we used the submissions passing at least one test and used abstract syntax trees to remove all user-generated variable names and function signatures, replacing them with random identifiers. Due to the nature of the AST-Rules baseline, this ablation did not affect the results for that approach. We ran all of the remaining approaches with the modified data.
% The goal of this prompt is to see whether the LLM model relies more on keywords (e.g. a variable called `counter') or the underlying structure of the code. As students may not be using descriptive variable names in their code, this modification may provide more realistic results.

% We evaluate the two core models, \textbf{AST} and \textbf{GPT-4o}, on both instructor solutions and student submissions. The evaluation on instructor solutions acts as a sanity check as the models were designed on this set of data. We evaluate the two LLM variants on student submissions only, as our main interest is observing the difference in performance from the LLM model.
% - we filter the dataset to get submissions that have all the plans in the instructor solution or have no plans at all
% - a correct prediction is each one of the plans
\section{Results}

\subsection{Exploratory Data Analysis}

The dataset had 10 classes including \textit{UNKNOWN} and the class distribution was imbalanced with most common being \textit{filterACollection} ($24.5\%$), and least common being \textit{linearSearching} ($3.2\%$). To address the imbalance, we computed per-class F1 scores and weighted averages in addition to overall accuracy scores.

% Before evaluating our classification models, we present an exploration of our dataset to capture potential imbalance/bias in the data distribution. Figure \ref{fig:plan-dist} shows the number of submissions that implement each plan, highlighting that some plans are used at a much higher rate compared to the average (\textit{filterACollection}, \textit{processAllItems}), and some plans are represented at lower rates (\textit{linearSearching}, \textit{multiWayBranching}). 
% This emphasizes the need for classification metrics that take class imbalances into account (e.g. weighted F1 scores) in addition to overall accuracy measures.

% \begin{figure}
%     \centering
%     \includegraphics[width=0.8\linewidth]{plan_distribution.png}
%      \caption{Number of submissions that practice each plan}
%     \label{fig:plan-dist}
% \end{figure}

Table \ref{table:eda-success} contains an analysis of the submissions by their test case success and by whether they use programming plans. Success on test cases is categorized into four groups as explained in Sec \ref{sec:dataset}. Programming plan use is reviewed under three categories: submissions that use the same set of plans as the instructor (\textit{Instructor Set}), submissions that use any other plan taught in the class (\textit{Class Set}), and submissions with no known plans (\textit{UNKNOWN}).
\begin{table}[b]
\centering
\caption{Performance on test cases by usage of programming plans}
% \begin{tabular}{lcccccc}
% \toprule
% % Submission Type & Total Submissions & \% Matching Instructor Plans & \% Expanded Instructor Plans & \% Missing Instructor Plans & \% Not Using Plans \\
% Success at Test Cases & Total & \% Matching & \% Expanded & \% Missing & \% Not Using Any \\
% % \multirow{2}{*}{Category} & \multirow{2}{*}{Total}  & \multicolumn{3}{c}{\% Passing} & \multirow{2}{*}{\% Not Compiling}\\
%                           % & & All Tests & Some Tests & No Tests & \\
% \midrule
% Passing All Tests & 692 & 77.17 & 4.77 & 13.44 & 4.62 \\
%  & 456 & 61.4 & 6.58 & 21.05 & 10.96 \\
% Passing No Tests & 234 & 62.39 & 3.85 & 17.09 & 16.67 \\
% Compilation Error & 234 & 63.68 & 4.7 & 12.82 & 18.8 \\
% \midrule
% Total Submissions & 1616 & 68.63 & 5.14 & 16.03 & 10.21 \\

% \bottomrule
% \end{tabular}

\begin{tabular*}{\textwidth}{@{\extracolsep{\fill}}lcccccc}
\toprule

\multirow{2}{*}{Success on Test Cases} & \multicolumn{3}{c}{\% Submissions Using Plans From} & \multirow{2}{*}{ \textbf{\#Total}} \\
           & Instructor Set & Class Set & UNKNOWN & \\

\midrule
Passing All Tests& 77.17 & 18.21 & 4.62  & \textbf{692}  \\
Passing Some Tests& 61.40 & 27.63 & 10.96  & \textbf{456}  \\
Passing No Tests& 62.39 & 20.94 & 16.67  & \textbf{234}  \\
Compilation Error& 63.68 & 17.52 & 18.80  & \textbf{234}  \\
\midrule
Overall& 68.63 & 21.16 & 10.21  & \textbf{1616}  \\

\bottomrule
\end{tabular*}
\label{table:eda-success}
\end{table}
We observed that a higher percentage of successful submissions use the same plans as instructor solutions ($\sim77\%$ vs $\sim62\%$), supporting our assumption that some students fail at these exercises due to a mistake in the plan selection stage. Similarly, we see that less than $5\%$ of successful submissions omit plans, and the percentage of submissions with no known plans increases as the success of submissions decreases, shown by almost $19\%$ of non-compiling submissions missing any plans from the class. While these results are not conclusive, they reinforce the idea that students who can select correct plans create correct submissions at higher rates, pointing to the potential benefits of plan-based feedback.

% \begin{table}[h!]
% \centering
% \caption{COMMENT OUT - Column-Wise Percentage Breakdown of Submissions by File (Transposed)}
% \begin{tabular*}{\textwidth}{@{\extracolsep{\fill}}lcccccc}
% \toprule
% \multirow{2}{*}{Submissions with} & \multicolumn{3}{c}{\% Submissions Passing} & \% Error at & \multirow{2}{*}{Total} \\
%                           & All Tests & Some Tests & No Tests & Compilation & \\
% % Category & \% Passing All & \% Passing Some & \% Passing No & \% Syntax Error & Total\\
% \midrule
% % Submissions with EXCLUDE & 56.86 & 43.14 & 0.0 & 0.0 & 51.0 \\
% Instructor Plans & 48.15 & 25.25 & 13.17 & 13.44 & 1109 \\
% Different Plans & 36.84 & 36.84 & 14.33 & 11.99 & 342 \\
% Not Using Plans & 19.39 & 30.3 & 23.64 & 26.67 & 165 \\
% \midrule
% Total Submissions & 692 & 456 & 234 & 234 & 1616 \\
% \bottomrule
% \end{tabular*}
% \end{table}

% \begin{table}[ht]
%     \centering
%     \begin{tabular}{ccccc}
%     \toprule
%          \textbf{Model} & \textbf{Instructor} & \multicolumn{3}{c}{\textbf{Student}} \\  \cmidrule(lr){3-5}
%          & & Overall & Correct & Partial \\\midrule
%          AST & 67\% & 65\% & 72\% & 50\% \\ \midrule
%          GPT4o & 94\% & 92\% & 95\% & 86\% \\
%          GPT4o-mini & --- & 86\% & 93\%  & 69\% \\
%          GPT4o-mini-masked & --- & 86\% & 92\% & 72\% \\
%          \bottomrule
%     \end{tabular}
%     \caption{Caption}
%     \label{tab:my_label}
% \end{table}

\subsection{Main Study: Success on Plan Classification}

% \begin{table}[ht]
% \small
%     \centering
%         \caption{Prediction accuracies for models for completely correct, partially correct, and all submissions.}
%     \begin{tabular}{cccc}
%     % \toprule
%          \textbf{Model} & \multicolumn{3}{c}{\textbf{Student Submissions}} \\ % \cmidrule(lr){2-4}
%          & Complete & Partial & \textit{All} \\%\midrule
%          AST & 67\% & 50\% & 60\%  \\ %\midrule
%          knn & 40\% & 35\% & 38\%  \\ %\midrule
        
%          GPT-4o & 85\% & 78\% & 82\% \\
%          GPT-4o-mini & 82\%  & 64\%  & 75\%\\

%          GPT-4o-ft & 83\% & 76\% & 80\% \\
%          GPT-4o-mini-ft & 86\%  & 79\%  & 83\%\\
         
%          GPT-4o-mini-masked & 82\% & 68\% & 77\% \\
%          % Deep-Coder-v2 & 79\%  & 53\%  & 69\%\\
%          % \bottomrule
%     \end{tabular}
%     \label{tab:accuracy_results}
% \end{table}
% Table \ref{tab:accuracy_results} shows the accuracy of each model on completely correct student submissions, partially correct student submissions, and overall. We see that AST approaches achieve 72\% accuracy on completely correct submissions, but their performance drops to 50\% on partially correct submissions, emphasizing their sensitivity to small syntax errors. On the other hand, GPT-4o achieves 95\% on completely correct and 86\% on partially correct submissions, outperforming the AST baseline by 23\% and 36\% respectively. 

Table \ref{tab4.2:all-metrics-overall} shows the overall classification accuracy of each approach, averaged over all submissions. We use three metrics for quantifying classification accuracy. Exact match ratio is a strict accuracy metric that measures the ratio of \textit{submissions that are correctly classified}, penalizing partial cases where the model produced an incorrect classification for one of the several plans involved in a single submission. Micro-F1 score provides a more lenient accuracy metric that measures the ratio of \textit{plans that are correctly classified}, providing credit to cases with partial success where the model classified at least one plan correctly in a submission with multiple plans.
Finally, the weighted F1 score provides a more nuanced view by calculating the F1 scores for each plan individually and averaging per-plan F1 scores weighted by the number of submissions including each plan to adjust for data imbalance. Micro-F1 is preferred when the performance on each plan is equally important, but weighted-F1 might be more accurate for estimating actual student experience if students are required to use some plans at a much higher frequency than others.

\begin{table}[b!]
\centering
\caption{Comparison of all approaches by three evaluation metrics}
\begin{tabular*}
{\textwidth}{@{\extracolsep{\fill}}lccc}
\toprule
\multirow{1}{*}{Approach} & Exact Match Ratio & Micro-F1 Score & Weighted-F1 Score \\
% & Ratio & (Micro) & (Weighted) \\ 
\midrule
Baseline & & & \\ 
\midrule
AST-Rules & 0.5259 & 0.5880 & 0.6035 \\ 
CodeBERT-kNN & 0.4965 & 0.5496 & 0.5315 \\ 
\midrule
With Prompting & & & \\ 
\midrule 
GPT-4o & 0.7157 & 0.7779 & 0.7647 \\ 
GPT-4o-mini & 0.6070 & 0.6721 & 0.7016 \\ 
\midrule
With Finetuning & & & \\ 
\midrule
GPT-4o & 0.7016 & 0.7713 & 0.7395 \\ 
GPT-4o-mini & 0.7157 & 0.7816 & 0.7442 \\ 
\bottomrule
\end{tabular*}
\label{tab4.2:all-metrics-overall}
\end{table}

LLM approaches outperform baseline approaches significantly in all metrics. 
We observed that both baseline approaches perform similarly, with KNN-Clustering performing slightly worse than the AST-Rules approach. LLM-based approaches provide remarkably better results even with no fine-tuning applied, with GPT-4o and GPT-4o-mini improving the baseline by $.20$ and $.10$ points in micro-F1 scores.
A series of Wilcoxon signed-rank tests with Bonferroni corrections indicated that the F1 scores for all GPT approaches were significantly different than both baselines ($p < .001$), with no significant difference between the baselines AST and KNN ($S=174138, p=.27$). If fine-tuning is not possible, GPT-4o with few-shot prompting could be a viable model for providing feedback.

Promisingly, fine-tuning GPT-4o-mini improves its performance by another $.10$ points, with the best approach by micro-F1 score overall being GPT-4o-mini ($.782$). Even with a relatively small dataset and a short post-training process, this small and cost-effective model can outperform the larger, state-of-the-art model.
We note that GPT-4o's performance is slightly hindered by fine-tuning, potentially implying that the larger models do not benefit from being fine-tuned on small datasets, but the difference is not significant ($S=10133.5, p=.61$). There were no significant differences between prompted 4o and fine-tuned 4o-mini ($S=9564, p=.33$) or between fine-tuned 4o and 4o-mini ($S=6432, p=.07$).

\begin{table}[t!]
\centering
\caption{Micro-F1 scores compared for four types of student submissions}
\begin{tabular*}{\textwidth}{@{\extracolsep{\fill}}lcccc}
\toprule
\multirow{3}{*}{Approach} & \multicolumn{4}{c}{Micro-F1 Score for Submissions} \\ & Passing All & Passing Some & Passing No & With Syntax \\ 
& Tests & Tests & Tests & Error \\
% \multirow{2}{*}{Approach} & Passing All & Passing Some & Passing No & Syntax Error \\ & Tests & Tests & Tests & \\ 
\midrule
Baseline & & & \\ 
\midrule
AST-Rules & 0.7419 & 0.5874 & 0.5325 & 0.1778 \\ 
CodeBERT-kNN & 0.6500 & 0.5138 & 0.4218 & 0.4561 \\ 
\midrule
With Prompting & & & \\ 
\midrule
GPT-4o & 0.8373 & 0.7597 & 0.6891 & 0.7258 \\ 
GPT-4o-mini & 0.8255 & 0.6031 & 0.4775 & 0.5310 \\ 
\midrule
With Finetuning & & & \\ 
\midrule
GPT-4o & 0.8320 & 0.7391 & 0.7371 & 0.6892 \\ 
GPT-4o-mini & 0.8410 & 0.7805 & 0.6970 & 0.6940 \\ 
\bottomrule
\end{tabular*}
\label{tab4.2:base-metrics}
\end{table}

We show that LLM approaches are especially valuable for providing feedback for code with errors in Table \ref{tab4.2:base-metrics}, where micro-F1 scores for each approach on four types of submissions are provided. These results show that all approaches achieve worse prediction accuracy on submissions with errors. However, we also see that the performance gap between the baselines and the LLM approaches widens in submissions with errors. While LLM's performance decreases by approximately $10\%$ in failing submissions, we see that the baseline models are almost unusable, experiencing drops in predictive performance that range from $30\%$ to $70\%$.
% Encouragingly, we see that the fine-tuned GPT-4o-mini results are better than or comparable to other approaches in all subsets of the data, showing that it could be a valid alternative to the more costly state-of-the-art model.

\begin{figure}[ht]
    \centering
\includegraphics[width=.9\linewidth]{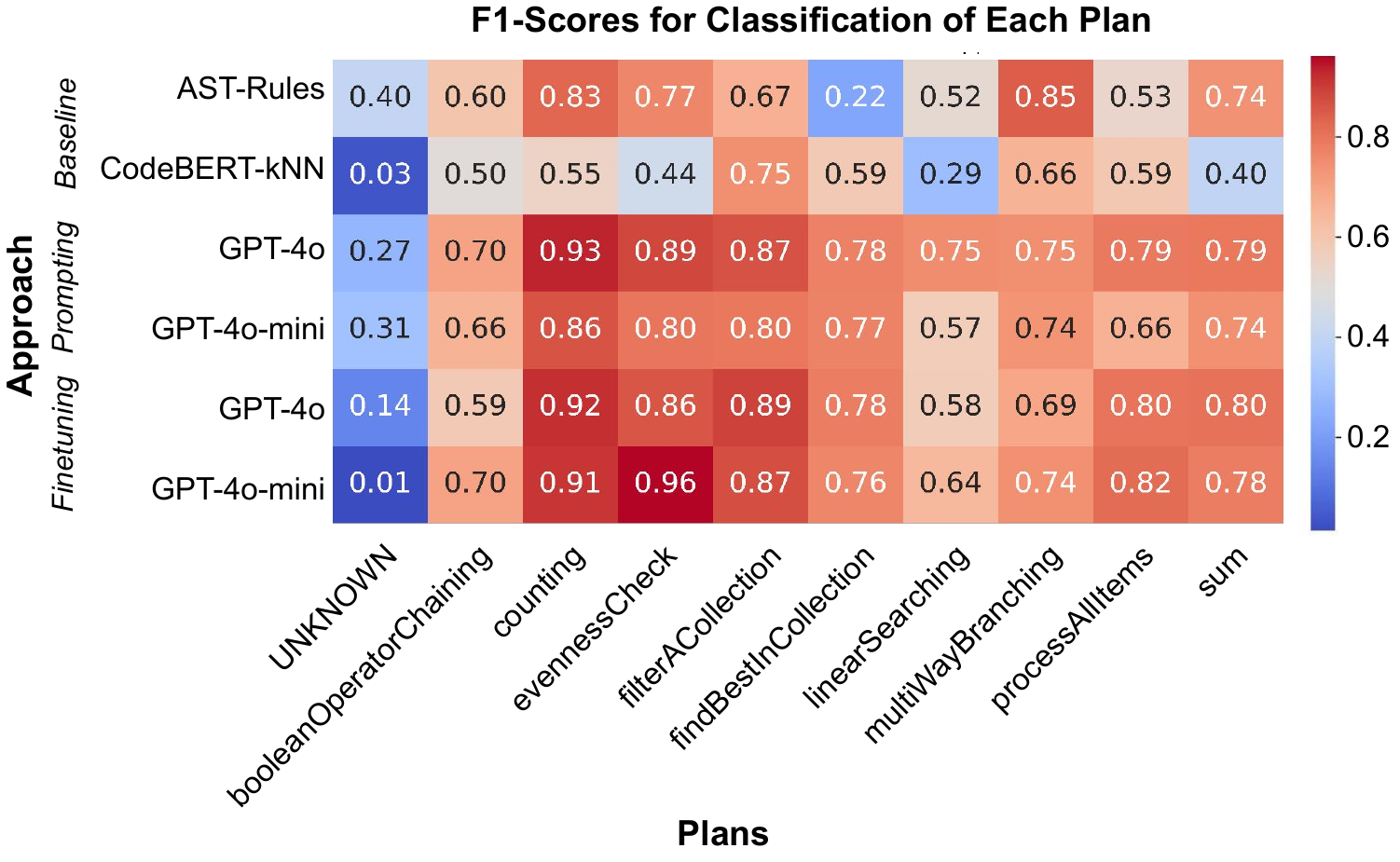}
    \caption{F1 scores for each of the ten programming plans. Models perform noticeably worse at classifying solutions that are labeled as UNKNOWN.}
    \label{fig:f1-heatmap}
\end{figure}

% \textbf{Per plan F1 scores}
We note that the model performance can vary by plan as shown in Figure \ref{fig:f1-heatmap}, with the F1 scores calculated for each plan. The heatmap highlights two important insights that are not obvious from the aggregate metrics. First, submissions with no known plans (represented with label \textit{UNKNOWN}) seem to be classified more poorly than other plans. This indicates that all our approaches are biased towards predicting \textit{any} plan rather than predicting the \textit{UNKNOWN} token. Second, fine-tuned models, especially fine-tuned GPT-4o-mini, perform especially worse at predicting \textit{UNKNOWN}s. Note that the fine-tuning is done on the instructor solutions, which do not include any datapoints with \textit{UNKNOWN} label. Thus, the fine-tuned models seem to be biased towards repeating labels from the data distribution rather than following the rules laid out in the prompt.
Otherwise, there are no notable exceptions to the trends observed in the previous tables. 

\subsection{Ablation Study: Impact of Obfuscation}

Table \ref{tab4.3:obf-metrics} shows the results of our ablation study, testing all models on completely/partially correct submissions with obfuscated variable names. For each approach, micro-F1 scores and the change from the original results are shown. Note that the results for the AST-Rules baselines are unchanged, as ASTs discard variable names during canonicalization.

\begin{table}[b!]
\centering
\caption{Micro-F1 scores for obfuscated data and differences from original scores}
\begin{tabular*}{\textwidth}{@{\extracolsep{\fill}}lcc}
\toprule
\multirow{2}{*}{Approach} & \multicolumn{2}{c}{Micro-F1 Score for Submissions} \\ & Passing All Tests ($\Delta$) & Passing Some Tests ($\Delta$) \\ 
\midrule
Baseline & & \\ 
\midrule
AST-Rules & 0.7419 (0.0000) & 0.5874 (0.0000) \\ 
CodeBERT-kNN & 0.5304 (-0.1197) & 0.3677 (-0.1460) \\ 
\midrule
With Prompting & & \\ 
\midrule
GPT-4o & 0.8419 (0.0046) & 0.7543 (-0.0054) \\ 
GPT-4o-mini & 0.8363 (0.0108) & 0.6504 (0.0472) \\ 
\midrule
With Finetuning & & \\ 
\midrule
GPT-4o & 0.8379 (0.0059) & 0.7576 (0.0184) \\ 
GPT-4o-mini & 0.8479 (0.0069) & 0.7813 (0.0008) \\ 
\bottomrule
\end{tabular*}
\label{tab4.3:obf-metrics}
\end{table}

LLM approaches seem to be robust to these changes on the surface-level features.
% We did not observe any meaningful changes on the classification performance of LLMs, with changes ranging between $-1\%$ and $+3\%$. 
However, we see that obfuscation hurts the performance of the CodeBERT-KNN baseline by around $20\%$, A Wilcoxon signed-rank test indicated that that the decrease for KNN was statistically significant ($S=18981.5, p<.001$), whereas no significant differences were observed for GPT models (for prompting: GPT-4o $S=2667.0, p=.97$, GPT-4o-mini $S=5106.5, p=.02$; for fine-tuning: GPT-4o $S=2,100, p=.07$, GPT-4o-mini $S=1466.5, p=.35$), suggesting that CodeBERT embeddings are sensitive to the surface-level features that the ablation study removes, as we initially hypothesized. 

\section{Discussion}

% Main study results - LLMs are better at this task
Large language models are far more accurate in providing feedback on programming plans compared to conventional baselines using graph representations or code similarity measures. We achieve remarkable micro-F1 scores by prompting larger models such as GPT-4o with a few example submissions, in line with earlier studies on few-shot prompting~\cite{wuMatchingExemplarNext2023}. In grading tasks where prior submission data is limited, LLM-based autograders provide a reasonable starting point.

% Finetuning results - small models
Promisingly for classroom deployments, the smaller GPT-4o-mini model achieved and surpassed the performance of the state-of-the-art model after being fine-tuned on the small subset of the data. GPT-4o-mini is less computationally expensive, more cost-effective, and could be deployed in real-time grading scenarios at $1/16$th of the cost of GPT-4o for the same performance. While we focused on GPT models, these findings also motivate fine-tuning open-source language models. There are some ethical and privacy concerns associated with using third-party APIs as part of a grading pipeline. Open-source models can be hosted on institution servers or even student devices, allowing the instructors or students to retain control of their data by processing them in secure environments.

In addition to overall higher performance, one main advantage of LLMs over baseline methods is to provide planning feedback on code submissions that are incorrect or incomplete (Table \ref{tab4.2:base-metrics}). Prior studies on LLM autograders have reported that LLMs are prone to generating correct code from incorrect student artifacts, such as traces of planning activities or handwritten pseudocode~\cite{riveraIterativeStudentProgram2024,10.1145/3657604.3662027}. In our work, we use this overcorrection tendency to provide high-level feedback on the structure the student \textit{intended} to implement by ignoring implementation errors. Thus, LLMs not only improve the accuracy of plan feedback on correct submissions compared to baselines, but also make it possible to generate feedback for submissions with syntactic or semantic errors.

% Preliminary study shows that plan mismatches are an important reason for errors
% Students' mistakes in the planning stage 
% % , particularly choosing an incorrect pattern/plan, 
% seem to be a considerable reason for incorrect submissions, even though they are not directly addressed in traditional autograders.
% Our exploratory analysis from Sec \ref{} shows that passing submissions tend to use programming plans as the instructor intended, and submissions missing programming plans tend to fail at compilation. While we cannot draw causal conclusions from this data, we can observe a pattern between utilizing programming plans correctly and passing the programming exercises. Therefore, providing any kind of feedback to help students develop their planning skills and programming plan use can be impactful in increasing their success on programming exercises.

% \authornote{teaching implications - feedback can improve motivation? Max took a stab here}
\textbf{Teaching implications.} There are multiple potential benefits to providing feedback on programming plans. First, our exploratory data analysis shows that when students' selected plans differ from the instructor's plans, submissions are more likely to be incorrect. Targeting the development of planning skills directly may improve students' programming exercise performance.
Furthermore, feedback that aims to help students identify plans could potentially lead to more reflection on the problem-solving process. The literature suggests that reflective feedback~\cite{abu2024impact} and feedback that nudges as opposed to providing direct next steps~\cite{zamprogno_nudge_2020} in autograding systems may improve students' performance and interaction with feedback. Additionally, immediate feedback may improve student performance and increase students' willingness to submit assignments~\cite{mitra_immediate_23}. 
In the programming context, recent work augmenting SQL feedback with hints generated by comparing model solutions and student queries saw students require fewer submissions to construct correct solutions~\cite{kleiner_sqlfeedback}, suggesting faster learning.

However, deployment of automatic feedback does not come without some risk, particularly when an autograder is \textit{incorrect}. One study investigating an NLP autograder for students' short descriptions of code found that false positives (that is, saying a student is correct when they are not) reduced student learning, potentially due to reducing reflection~\cite{li_hsu_wrong_23}. However, false negatives were not harmful, as students were more able to reflect upon the feedback.
Given that our plan feedback approaches can get relatively low F1 scores on some classes (e.g. \textit{UNKNOWN} token), plan feedback can be presented in ways to encourage reflection rather than to generate final grades. 
% Given our approach to identifying programming plans struggles with the \textit{UNKNOWN} token, future work may wish to investigate the degree to which incorrect plan detection impacts student learning and responses to feedback.
% Given that automatic feedback that encourages reflection is particularly useful and the relatively low accuracy of even our best approaches, around 70-75\%, future work should investigate the framing of programming plan feedback. 
For example,
% rather than explicitly identifying a plan by name, and risking a falsely identified plan name confusing students, 
feedback could present students with worked examples that incorporate plans the system \textit{thinks} the student was trying to use. 
% Not only would providing a worked example constitute feedback that is potentially more reflective than a pattern name, the risk may be lower: 
At worst, a mismatch between a worked example and a plan a student was using may be ignored by the student. %Rather than incorrectly providing credit to a student for a plan they could not identify correctly, they may at worst receive a worked example they cannot connect to the problem and choose to ignore. 

\textbf{Limitations and future work.} 
Our work provides a first step for feedback on planning in code-writing problems by identifying programming plans in short programs. Future work can explore ways to generalize this approach to larger programming projects, where students may need to modify and combine multiple plans.
Furthermore, evaluating this detection technique in a real-time environment with students can yield a greater understanding of how getting feedback on the problem-solving process can shape the student experience.

Due to the rapid nature of LLM research, we left recent reasoning models out of our scope. We found that these models with Chain-of-Thought generation increased inference time substantially in preliminary experiments, with more than 30 seconds per submission (compared to <1 second in our models). Thus, these models may not be appropriate for real-time grading at large scale.

One exciting finding from our ablation study was the implication that LLMs classify submissions based on their structure and not on surface-level context cues like keywords. Therefore, future work could explore the use of LLMs to process code for analyzing structure and subgoal-level information. Moreover, the strength of LLMs in pattern recognition can motivate their application in non-programming domains where students practice selecting and applying patterns, including proof techniques in math~\cite{lee2016students}, system analysis problems in physics~\cite{hewson1984use}, or schema acquisition for language learning~\cite{carrell1984schema}.

% \authornote{can mention other pattern based domains as potential future work (math, proofs etc) (due to LLMs strenghts in pattern recognition)}
% Also, while we present a baseline using handcrafted AST analysis rules, hybrid approaches that use machine learning techniques on information from ASTs may combine the strengths of both approaches into one. Finally, evaluating this detection technique in a real-time environment with students can yield a greater understanding of how getting feedback on various levels of programming skills can shape the student experience.

% One opportunity for future work is adapting smaller language models or open-source models for this task. By fine-tuning these smaller models, it may be possible to obtain results on par or better than GPT-4o. Moreover, these models can run locally on personal devices, making this technique accessible to a larger group of students. 
% A limitation that may be explored in future work is the extent to which AST-based approaches can be developed. 

% \section{Limitations and Future Work}
% - only on short examples
% - This catches both correct and incorrect submissions
% - This could provide worked examples to help students fix their code, if they have the correct plan
% - This could prompt students with high level questions if they haven't got the right plan yet
% - This could provide students with partial credit

\section{Conclusion}
In this work, we propose a framework for generating high-level planning feedback for open-ended programming exercises. By analyzing data from a CS1 course, we show that large language models can provide this feedback at higher accuracy compared to traditional code analysis methods. Moreover, even small models can be fine-tuned to provide accurate feedback at a lower cost. This approach can support students as they develop intermediate planning skills in programming problems, as well as students in other domains where recognizing and applying common patterns plays an important role.

\begin{credits}
% \subsubsection{\ackname} Acknowledgments

% \subsubsection{\discintname}
% Conflict of Interest
\end{credits}
\bibliographystyle{splncs04}
\bibliography{references-doi, references-archive}
\end{document}